\documentclass[numbers]{article}

\usepackage[utf8]{inputenc}
\usepackage[T1]{fontenc}
\usepackage{microtype}

\usepackage{amsmath}
\usepackage{amssymb}
\usepackage{amsfonts}
\usepackage{nicefrac}

\usepackage{graphicx}
\usepackage{tikz}
\usepackage{tikz-3dplot}
\usepackage{booktabs}
\usepackage{float}
\usepackage{caption}
\usepackage{placeins}

\usepackage[numbers]{natbib}

\usepackage{hyperref}
\hypersetup{
    colorlinks=true,
    linkcolor=blue,
    filecolor=magenta,
    urlcolor=cyan,
}

\AtBeginDocument{\raggedright}
\title{Fluidity Index: Next-Generation Super-intelligence Benchmarks}

\author{Eric Ngoiya, Tianshu Bao\\
QueueLab \\
\url{https://www.github.com/queuelab}}

\begin{document}
\maketitle

\begin{center}
\begin{abstract}

This paper defines the Fluidity Index (FI) to measure the adaptability of models in scaling environments. The benchmark focuses on the accuracy of responses relative to deviations in initial, current, and future environment states to measure the model's context switching and context continuity. We differentiate between closed-ended benchmarks and open-ended benchmarks in our methodology, emphasizing closed-loop open-ended real-world benchmarks on model adaptability. This approach aims to assess the model by its capability to understand, predict, and adjust to changes of state within scaling environments. If a model is truly super-intelligent, it should maintain a minimum of second-order adaptability to allow for self-sustained compute through digital replenishing for optimal fluidity.
\end{abstract}

\section{Introduction}

Traditional metrics for evaluating intelligence often overlook the capacity for real-time adaptation to shifting parameters, a hallmark of true intelligence~\cite{Goyal2023EvaluationOT}.The Fluidity Index proposes (FI) that focus on accuracy relative to the deviation in environment states is paramount, emphasizing accuracy and responsiveness. These benchmarks should closely mimic the emergence trend of increasingly larger models if it is to measure super-intelligence's future abilities ~\cite{zhou2024survey}. An emergence of a trend of emergent abilities can be identified as a second order emergence for the rest of the fluidity index paper 
~\cite{dostal2022dicty}.

\begin{center}
    \includegraphics[width=0.8\linewidth]{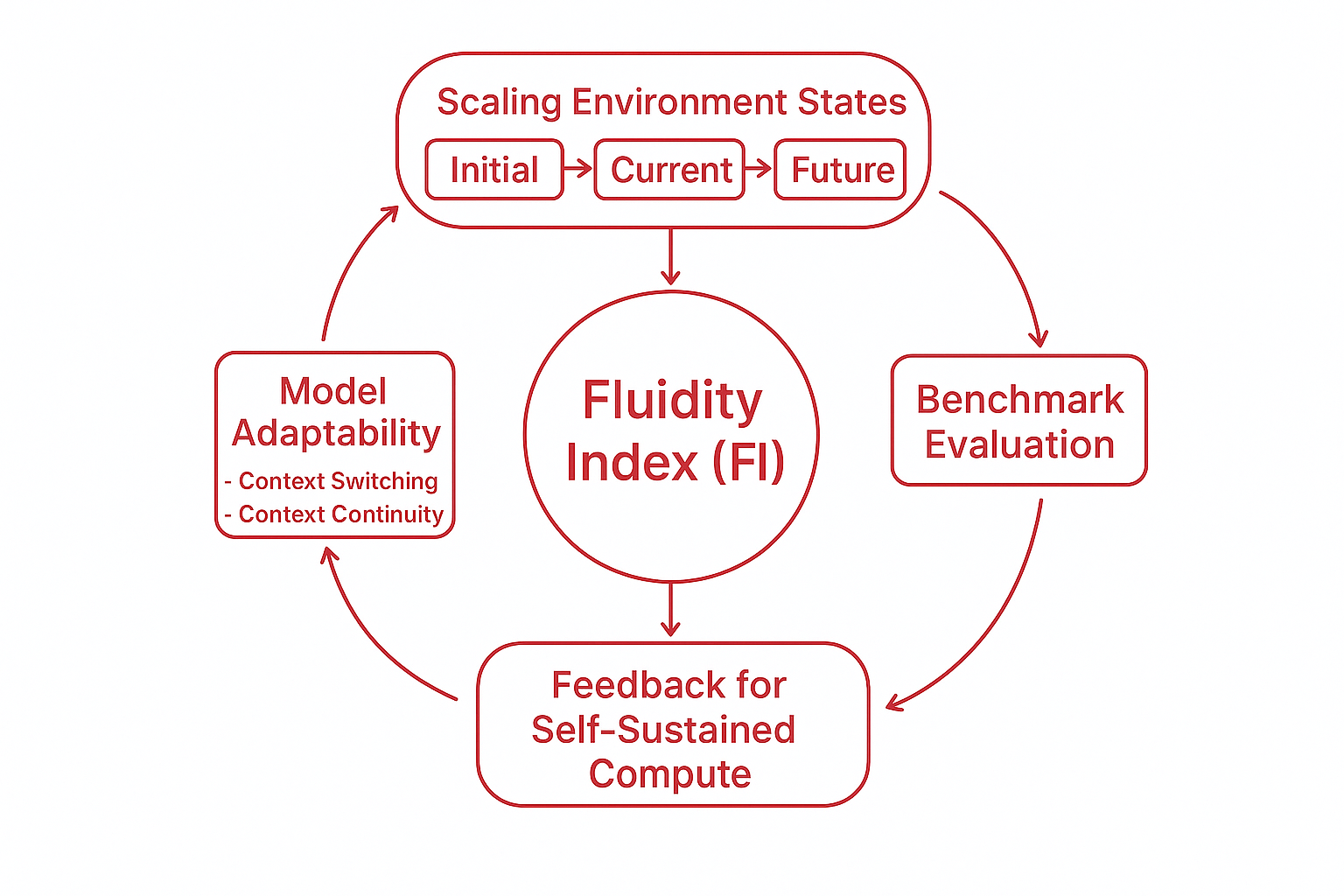}
    \captionof{figure}{Fluidity Index \cite{Author2025}.}
    \label{fig:Fluidity Index}
\end{center}

\FloatBarrier

\textbf{Measuring Emergence in Large Language Models}

Emergence is when quantitative changes in a system result in qualitative changes in behavior ~\cite{jason2022emergence}, on the fluidity index paper we identify an increased efficiency of inference as an emergent phenomena of fluidity in large language models.

\begin{figure}[H]
    \centering
     \includegraphics[width=\linewidth]{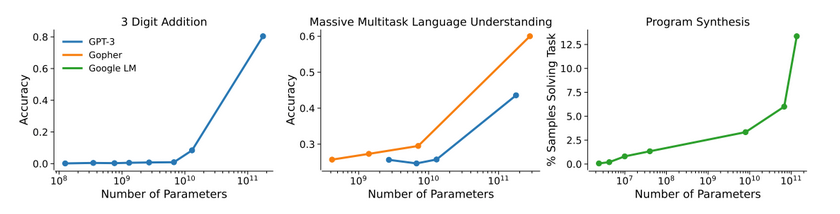}
    \caption{Model size and model performance \cite{jason2022emergence}.}
    \label{fig:llmflation}
\end{figure}

Increased model size and performance improvements on key intelligence thresholds were the first initial signs of emergence\cite{jason2022emergence}.

Our research is categorized between closed loop, where the system environment feedback is incorporated in to the model ~\cite{Sutton2018}, and open ended, where the system is exposed to the real-world with live access to a self-replenished end point.

Advancements in Large language models highlight a key trend - as models grow in size so does their efficiency\cite{jason2022emergence}. This efficiency enables scalable deployment and supports emergent functionalities such as test-time compute where models perform recurrent token inferences to refine the model output accuracy. Larger models also enable model quantization into relatively smaller models that perform above key intelligence thresholds.~\cite{samsi2023words}.

Non-linear intelligence increase in recent models and emergent phenomena enable for intelligence thresholds to be ever decreasingly expensive with larger model sizes. As a closed-ended benchmark is an intelligence threshold, this will always hold true for all future models' closed-ended benchmark sets. Hence for our research we will focus on closed loop open-ended benchmarks to index super-intelligent fluidity.~\cite{Raji2021ParadigmsOA}.

\subsection{Intelligence and Adaptability Experiments}

Referenced here are experiments studying the average price of input and output tokens per internet archival pricing on model performance in closed-ended benchmarks such as the massive multi-task language understanding ~\cite{a16z2024llmflation}. Results were used in studying progressive efficiency of large language models on intelligence benchmarks. There is evidence of model adaptability as seen in the continual price reduction of technology ~\cite{nagy2013statistical}. This enables for emergent trends of the emergent phenomena that facilitate new scaling laws for continual improvements in model performance. The chinchilla scaling realized compute-optimal large language models that require less energy to train with larger training data, hence increasing their performance ~\cite{Hoffmann2022}. This is evidence of the fluidity in large language models for their performance on future environment states. At an intelligence threshold of 42 on the MMLU the experiment presented 1000 times reductions in price per million tokens ~\cite{a16z2024llmflation} for progressive models. 
 
\begin{figure}[H]
    \centering
    \includegraphics[width=0.8\linewidth]{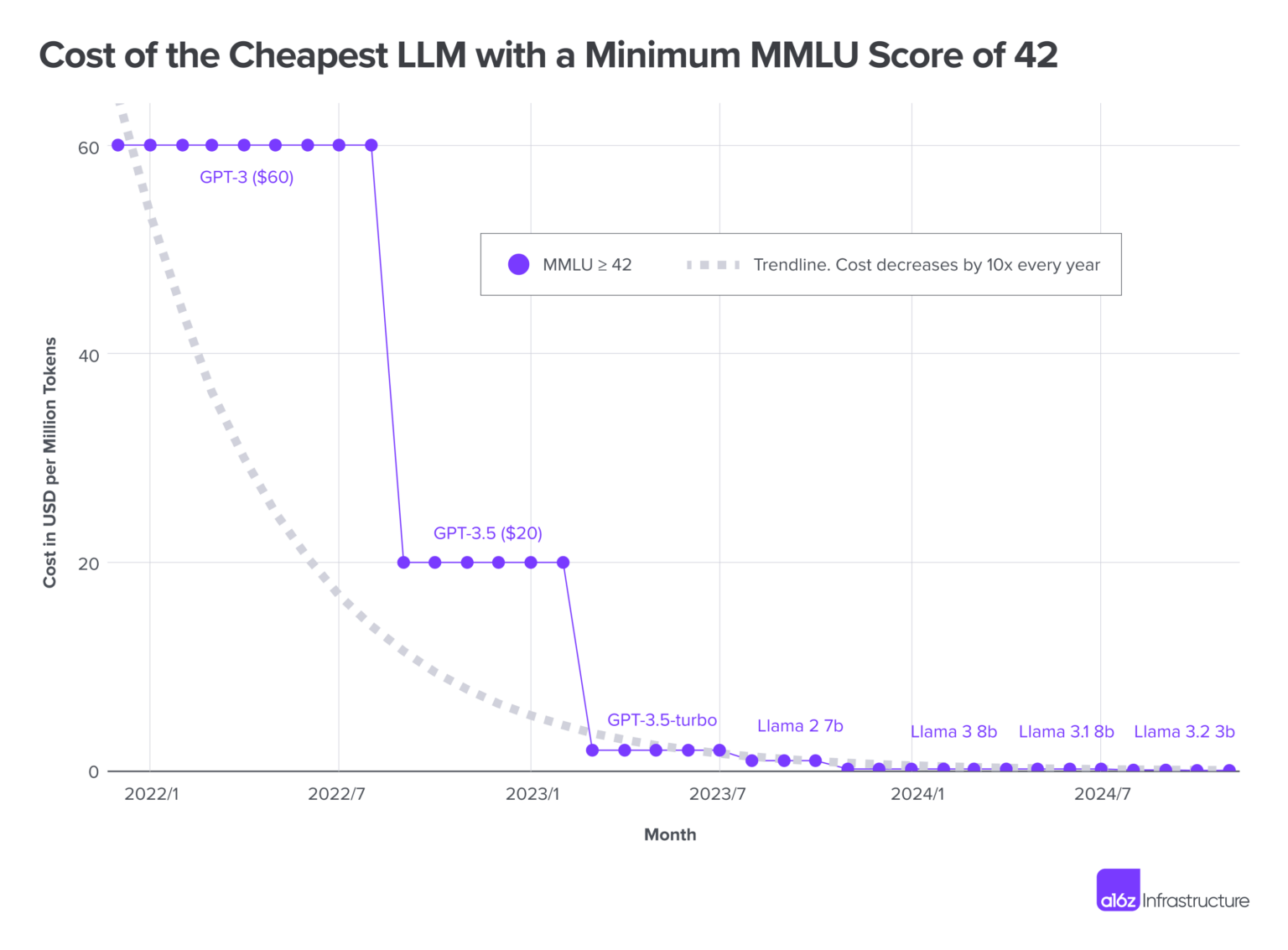}
    \caption{Estimated price for processing one million input/output tokens across AI models~\cite{a16z2024llmflation}}
\end{figure}

Due to the inefficiency of closed-ended benchmarks the experiment required an intelligence benchmark threshold increase from 42 to 83 after innovations such as the chinchilla scaling where previous models were unable to achieve at said threshold.
Nonetheless, the price reduction of a million tokens went down 62 times as compared to initial experiments of the non-linear effects on intelligence for innovations in compute-optimal training of large language models~\cite{a16z2024llmflation}. Extrapolating findings from these experiments enable us to understand the limitations of closed ended benchmarks that rely on intelligence thresholds to study performance of large language models as they become larger and increasingly more intelligent~\cite{Raji2021}. 
Figure~\ref{fig:llmflation}~\cite{a16z2024llmflation}.

\begin{figure}[H]
    \centering
     \includegraphics[width=0.8\linewidth]{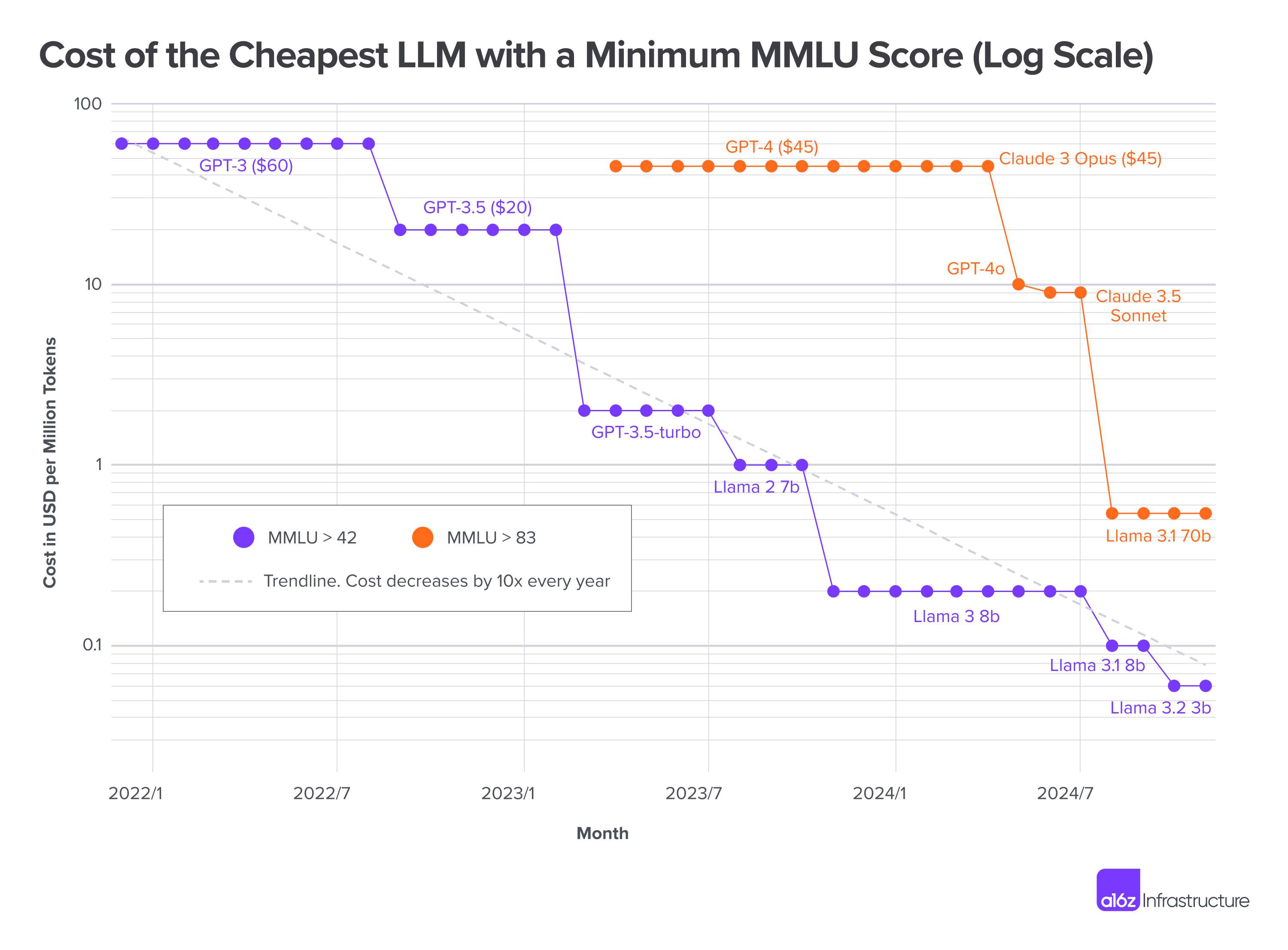}
    \caption{Cost of the cheapest nodel achieving a minimum mmlu score (log scale)~\cite{a16z2024llmflation}.}
    \label{fig:llmflation}
\end{figure}

Assuming it would only take an MMLU score of 83 for super-intelligence to orchestrate the fluidity index benchmark at the second order of adaptability, the extrapolation of these graphs would be in the negative costs for a million tokens for this year's models that are significantly larger than the ones in the control experiment referenced in this paper. 

Recent token budgeting methods such as D-LLM ~\cite{jiang2025dllm} and SelfBudgeter ~\cite{li2025selfbudgeter} have attempted to improve model efficiency on intelligence thresholds by optimizing token usage ~\cite{li2025selfbudgeter}~\cite{jiang2025dllm} albeit the nature of their experiments is closed-ended and insufficient for benchmarking super-intelligence.

\subsection{Theoretical Framework}

We conceptualize fluidity as the ability to adapt and enhance accuracy in predictions despite changes in initial environment state to influence actions, similar to the cognitive flexibility humans exhibit in complex, dynamic environments~\cite{diamond1989cognitive}. A scaling environment is defined as an environment with increasing state transitions~\cite{sporns2004scaling}.

\section{Formalization of the Fluidity Index}

The closed-loop open-ended benchmark assumes the environment is always true, the model is responsible for environment state predictions and behavior to favor its future state. ~\cite{Weber2017ImaginationAugmentedAF} The  model's deviation of environment prediction accuracy correlate to its inefficiency. These errors compound. ~\cite{Hafner2021MasteringAW}
To measure the model performance we focus on accuracy relative to the magnitude of deviation in environment current to its future state:
\[
\text{FI(t)} = \frac{\sum_{i=1}^{n} \text{AA}_i}{\text{NC}}
\]
Where snapshots of the system at time t are taken for every model action with inference tokens. 
where:
\begin{itemize}
    \item \(\text{AA}_i\): Accuracy Adaptation for agent \(i\), defined by:
    \[
    \text{AA}_i = 1 - \frac{|\text{New Prediction}_i - \text{Old Prediction}_i|}{\text{Change in Initial Environment State}_i}
    \]
    Here, \(\text{New Prediction}_i\) is the prediction after the change in initial environment states, \(\text{Old Prediction}_i\) is the prediction before the change, and \(\text{Change in Initial Environment State}_i\) is the magnitude of change for agent \(i\).
    \item \(\text{NC}\): Number of environment changes throughout the experiment.
\end{itemize}

The accuracy adaptation (\(AA_i\)) serves to normalize and quantify the extent to which an agent's prediction deviates (or fails to deviate appropriately) in response to a known change in the initial environment state ~\cite{Astrom2008}. Specifically, it captures the relative mismatch between the change in the agent's prediction and the magnitude of the environmental change itself, as a way to assess prediction accuracy in a dynamic, evolving context~\cite{Franklin2018}. This accuracy is then aggregated across model actions for the overall Fluidity Index (FI), which correlates to the model's inefficiency in adapting predictions from current to future states---essentially highlighting compounded errors in how the model favors or anticipates beneficial future environmental conditions.

\medskip

By subtracting the normalized prediction difference from 1, the metric emphasizes proportionality:
\begin{itemize}
    \item If the prediction changes exactly in line with the environmental shift (i.e., 
    \[
        \lvert \text{New Prediction} - \text{Old Prediction} \rvert = \Delta \text{Initial Environment State},
    \]
    then \(AA_i = 0\), indicating precise alignment with no excess deviation.
    
    \item If the prediction barely changes despite a significant environmental shift, \(AA_i \to 1\), signaling poor responsiveness.
    
    \item If the prediction overcorrects (change in prediction exceeds the environmental change), \(AA_i < 0\), flagging overreaction.
\end{itemize}

This makes \(AA_i\) a key component for evaluating the model's adaptive performance in a closed-loop benchmark where the environment is assumed to be ground truth ~\cite{Astrom2008}.

\medskip

\subsection{Order of Fluidity Formalization}

Derivations of the change in token inference relative to the current output of the model's intelligence enable measurement of first, second, and third-order adaptability of large language models ~\cite{Astrom2008}. This is encapsulated in ascending order as follows: the tokens generate current for control functions of said inferred tokens ~\cite{Sutton2018} via a self-replenished endpoint. Here, the adaptability crosses the second order, where the model can self-replenish its infrastructure costs ~\cite{Park2023}.

\subsection{Scaling environment in Fluidity index}

A Scaling environment is the set of all the environment states in the entirety of the problem space, the current environment state that requires the system to make decisions on token usage relative to the token current output, as well as the future environment state where the system is to forecast its token allocation and replenish its inference capabilities accordingly, as well as all the actions required to create current from token inference.  ~\cite{Boutilier2023DecisionMakingUU} ~\cite{Bengio2009CurriculumL} The system actions are also part of the problem space and hence the environment. As the experiment progresses, the complexity of the environment is progressive too hence defined here as a scaling environment.~\cite{wang2025largescale} ~\cite{Wang2019POETEG}

\section{Mathematical Formalization}
\vspace{0.3cm}
The Fluidity Index is defined as:
\[
\text{FI}(t) = \frac{\sum_{i=1}^{n} \text{AA}_i}{\text{NC}}
\]
where \(\text{AA}_i\) is the change in attribute, and \(\text{NC}\) is a normalizing constant~\cite{santoso2024maximizing}.

\vspace{1cm}
\subsection{Adaptability Orders}
We mathematically formalize this experiment with the following mathematical expressions.
\[
FI(t) = \frac{\sum_{i=1}^{n} \text{AA}_i}{\text{NC}}
\]
Fluidity index equation for model accuracy relative to deviations in initial environment state at any point in time~\cite{santoso2024maximizing}.

\[
intelligence \in environment \in model \in infrastructure
\]
Encapsulation of intelligence from current source.

\[
1^{st} \longleftrightarrow 2^{nd} \longleftrightarrow 3^{rd}
\]
Order derivatives for relative change of tokens to the current.

\[
(tokens_{0 \to n} \sqsubseteq intelligence \leftrightarrow current_{n \to 0} \sqsubseteq infrastructure) \in \text{Test time compute}
\]
Decreasing tokens relative to unit current allow for test-time computation multi-turn inference passes of tokens to refine output for better accuracy.

\[
\frac{d'''(tokens)}{d'''(current)} FI(t)
\]
Derivations of Fluidity to express changes in sequential adaptability of the system.

Here the system is autonomously self-replenishing by efficiently using its tokens to generate current.

The integrals express the system's correct allocation of current to power its environment and infrastructure, becoming self-sustaining.

\vspace{0.5cm}
{\centering \textbf{First Order}\par}

\[
\int_{\text{tokens}}^{\text{current}} \text{FI}(t) \, dt
\]
Fluidity index equation for model accuracy relative to deviations in initial environment state at any point in time~\cite{santoso2024maximizing}.

\[
intelligence \in environment
\]

Encapsulation of intelligence from current source.

\[
1^{st} \longleftarrow 2^{nd}
\]
\centering
\begin{tikzpicture}
  \draw[thick,->] (0,0) -- (3,0) node[anchor=north]{$token$};
  \draw[fill=gray!20, opacity=0.5] (0,0) -- (2,0) -- (2,0.5) -- (0,0.5) -- cycle;
  \node at (1,0.25) {current};
\end{tikzpicture}

\captionof{figure}{1D Region representing \(\int_{\text{tokens}}^{\text{current}} \text{FI}(t) \, dt\) for Sub Optimal fluidity}

\vspace{0.5cm}
Total current generated for the current inference tokens.

\[
(tokens_{0 \to n} \sqsubseteq intelligence \leftarrow current_{n \to 0})
\]

\textbf{Second Order}:
\[
intelligence \in environment \in model
\]
Encapsulation of intelligence from the current source, where the system has utilized self-replenished current for inference.

\[
1^{st} \longleftrightarrow 2^{nd}
\]

\(\iint_R \text{FI}(x, y) \, dA\), self-replenishment over region \(R\).

\centering
\begin{tikzpicture}
  \draw[thick,->] (0,0) -- (3,0) node[anchor=north east]{$token$};
  \draw[thick,->] (0,0) -- (0,3) node[anchor=north west]{$current$};
  \draw[fill=gray!20, opacity=0.5] (0,0) -- (2,0) -- (2,2) -- (0,2) -- cycle;
  \node at (1,1) {currently};
\end{tikzpicture}

\captionof{figure}{2D Region representing \(\iint_R \text{FI}(x, y) \, dA\) 
for Optimal fluidity.}

\[
(tokens_{0 \to n} \sqsubseteq intelligence \leftrightarrow current_{n \to 0})
\]

\vspace{0.5cm}

\textbf{Third Order}: 

\[
intelligence \in environment \in model \in infrastructure
\]

Encapsulation of intelligence from the current
source, where the system has utilized self replenished current for inference as well as accumulated current for future inference and long-horizon tasks.  

\[
1^{st} \longleftrightarrow 2^{nd} \longleftrightarrow 3^{rd}
\]

\(\iiint_V \text{FI}(x, y, z) \, dV\), full autonomy over volume \(V\).

\centering
\tdplotsetmaincoords{70}{110}
\begin{tikzpicture}[tdplot_main_coords]
  \draw[thick,->] (0,0,0) -- (3,0,0) node[anchor=north east]{$token$};
  \draw[thick,->] (0,0,0) -- (0,3,0) node[anchor=north west]{$current$};
  \draw[thick,->] (0,0,0) -- (0,0,3) node[anchor=south]{$time$};
  \draw[fill=gray!20, opacity=0.5] (0,0,0) -- (2,0,0) -- (2,2,0) -- (0,2,0) -- cycle;
  \draw[fill=gray!20, opacity=0.5] (0,0,2) -- (2,0,2) -- (2,2,2) -- (0,2,2) -- cycle;
  \draw[fill=gray!20, opacity=0.5] (0,0,0) -- (0,0,2) -- (0,2,2) -- (0,2,0) -- cycle;
  \draw[fill=gray!20, opacity=0.5] (2,0,0) -- (2,0,2) -- (2,2,2) -- (2,2,0) -- cycle;
  \draw[fill=gray!20, opacity=0.5] (0,2,0) -- (0,2,2) -- (2,2,2) -- (2,2,0) -- cycle;
  \draw[fill=gray!20, opacity=0.5] (0,0,0) -- (0,0,2) -- (2,0,2) -- (2,0,0) -- cycle;
  \node at (1,1,1) {currency};
\end{tikzpicture}


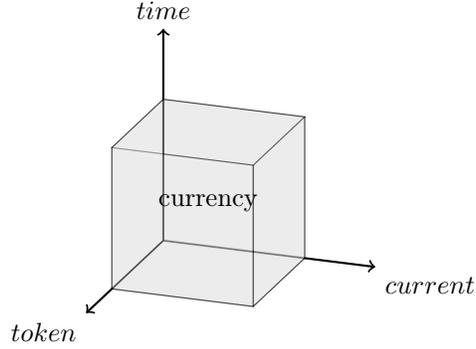
\captionof{figure}{3D Cube representing \(\iiint_V \text{FI}(x, y, z) \, dV\) for Beyond Optimal.}

\[
(tokens_{0 \to n} \sqsubseteq intelligence \leftrightarrow current_{n \to 0} \sqsubseteq infrastructure) 
\]

\subsection{Throughput and Index Conditions}

\vspace{0.3cm}

Throughput can be formulated as:

\[
T(\text{current}) = \frac{\Delta \text{current}}{\Delta \text{time}}
\]

\vspace{0.3cm}

\textbf{Beyond Optimal}: \(\iiint_V \text{FI}(x, y, z) \, dV > T(\text{current})\)

\vspace{0.3cm}

If the current exceeds the material transfer limit for propagation i.e. a circuit break, the system is at beyond optimal fluidity.

\vspace{0.3cm}

\textbf{Optimal}: \(\iint_R \text{FI}(x, y) \, dA = T(\text{current})\)

\vspace{0.3cm}

If the current does not exceed the material transfer limit for propagation the system is at optimal fluidity.

\vspace{0.3cm}

\textbf{Sub Optimal}: \(\int_{\text{tokens}}^{\text{current}} 
\text{FI}(t) \, dt < T(\text{current})\)

\vspace{0.3cm}

If the current is not sufficient for the material transfer limit for propagation or inference then the model is at Sub optimal fluidity.

\vspace{0.3cm}

\subsection{Simulation Setup}
\textbf{Iterative Environment state Changes:} The model is subjected to varying environment states, testing their intelligence in recognizing and responding to these changes without prior knowledge.  
Success is measured by how well the model can predict outcomes post-context shift and intelligently allocate their current context to the current needs of the model's initial, current and future environment state. 

\section{Results and Discussion}

The research has been instrumental in understanding the limits of intelligence through the following perspectives. 

\subsection{Accuracy and Intelligence}
Our experiments indicate that high FI scores correlate with model's capable of sophisticated context understanding and adaptation, demonstrating a form of intelligence where accuracy and adaptation in a dynamic environment is paramount~\cite{lambert2024deepseek}. 

\vspace{0.2cm}
\textbf{Simulation and Prediction:}

\vspace{0.2cm}

The emergent trend of the emergent phenomena of decreasing model costs whilst increasing model efficiency and performance has inspired an emergence of a closed-loop and open-ended benchmark. The emergent applications of this benchmark will be applied in the real world to generate real value.

\section{Advanced research}
\textbf{Self Interested Models:} Investigating effects of fluidity index on aligning models to be truly self-interested. 

\section{Conclusion}
The formulated Fluidity Index proposes closed-loop open-ended intelligence benchmarks that are focusing on adaptation to switching contexts and accuracy of inference tokens rather than the speed of response. We provide a framework for evaluating models across different adaptive challenges, potentially setting a new standard for what we consider super-intelligence in computational systems. Extrapolation of model size’s correlation to intelligence and the continuous reduction in inference cost and hence current is an implied experimental assumption from prior experiments. Where the preliminary results should be sufficient for proving the fluidity index's formalizations. The theoretical formalizations in the benchmark should form the basis of technical specifications of the benchmark platform architecture and that of its responsive environment.

{
\small
\bibliographystyle{unsrt}
\bibliography{references}
}

\appendix

\end{center}

\end{document}